\documentclass[11pt]{article}

\usepackage[utf8]{inputenc}
\usepackage{acl2014}
\usepackage{times}
\usepackage{url}
\usepackage{latexsym}
\usepackage{etoolbox}

\usepackage{graphicx}
\usepackage{hyperref}
\usepackage{url}
\usepackage{amsmath}
\usepackage{amssymb}
\usepackage{subfigure}
\usepackage{color}
\usepackage{caption}
\usepackage{multirow}
\usepackage{comment}
\usepackage{pdflscape}

\newcommand{\bmx}[0]{\begin{bmatrix}}
\newcommand{\emx}[0]{\end{bmatrix}}
\newcommand{\qt}[1]{\left<#1\right>}

\newcommand{\vect}[1]{\mathbf{#1}}
\newcommand{\vects}[1]{\boldsymbol{#1}}
\newcommand{\matr}[1]{\mathbf{#1}}

\newcommand{\vc}[0]{\vect{c}}
\newcommand{\vh}[0]{\vect{h}}

\newcommand{\vx}[0]{\vect{x}}
\newcommand{\vw}[0]{\vect{w}}
\newcommand{\vs}[0]{\vect{s}}
\newcommand{\vf}[0]{\vect{f}}
\newcommand{\ve}[0]{\vect{e}}
\newcommand{\vy}[0]{\vect{y}}
\newcommand{\vg}[0]{\vect{g}}
\newcommand{\vr}[0]{\vect{r}}

\newcommand{\mO}[0]{\matr{O}}
\newcommand{\mC}[0]{\matr{C}}
\newcommand{\mW}[0]{\matr{W}}
\newcommand{\mG}[0]{\matr{G}}

\newcommand{\mU}[0]{\matr{U}}
\newcommand{\mV}[0]{\matr{V}}

\newcommand{\TT}[0]{{\vects{\theta}}}

\newcommand{\RR}[0]{\mathbb{R}}

\newtoggle{arxiv}
\toggletrue{arxiv}

\graphicspath{ {./figures/} }

\title{Learning Phrase Representations using RNN Encoder--Decoder \\ for
Statistical Machine Translation}
\author{
    Kyunghyun Cho \\ 
    {\bf Bart van Merri\"enboer~ ~Caglar Gulcehre}\\
    Universit\'e de Montr\'eal \\
  {\tt \small firstname.lastname@umontreal.ca} \\
  \And
  ~\\{\bf Dzmitry Bahdanau} \\
    Jacobs University, Germany \\
  {\tt \small d.bahdanau@jacobs-university.de} \\
  \AND
    Fethi Bougares~~~~~Holger Schwenk \\
    Universit\'e du Maine, France \\
  {\tt \small firstname.lastname@lium.univ-lemans.fr} \\
  \And
    Yoshua Bengio \\
    Universit\'e de Montr\'eal, CIFAR Senior Fellow \\
  {\tt \small find.me@on.the.web} \\}

\date{}

\begin{document}
\maketitle

\begin{abstract}
    In this paper, we propose a novel neural network model called RNN
    Encoder--Decoder that consists of two recurrent neural networks (RNN). One
    RNN encodes a sequence of symbols into a fixed-length vector representation,
    and the other decodes the representation into another sequence of symbols.
    The encoder and decoder of the proposed model are jointly trained to
    maximize the conditional probability of a target sequence given a source
    sequence. The performance of a statistical machine translation system is
    empirically found to improve by using the conditional probabilities of
    phrase pairs computed by the RNN Encoder--Decoder as an additional feature
    in the existing log-linear model. Qualitatively, we show that the proposed
    model learns a semantically and syntactically meaningful representation of
    linguistic phrases.
\end{abstract}

\section{Introduction}

Deep neural networks have shown great success in various applications such as
objection recognition (see, e.g., \mbox{\cite{Krizhevsky-2012}}) and speech
recognition (see, e.g., \mbox{\cite{Dahl2012}}). Furthermore, many recent works
showed that neural networks can be successfully used in a number of tasks in
natural language processing (NLP). These include, but are not limited to, language
modeling~\mbox{\cite{Bengio2003lm}}, 
paraphrase detection~\mbox{\cite{SocherEtAl2011:PoolRAE}} and word embedding
extraction~\mbox{\cite{Mikolov2013}}. In the field of statistical machine
translation (SMT), deep neural networks have begun to show promising results.
\mbox{\cite{Schwenk2012}} summarizes a successful usage of feedforward neural
networks in the framework of phrase-based SMT system.

Along this line of research on using neural networks for SMT, this paper
focuses on a novel neural network architecture that can be used as a part
of the conventional phrase-based SMT system. The proposed neural network
architecture, which we will refer to as an \textit{RNN Encoder--Decoder},
consists of two recurrent neural networks (RNN) that act as an encoder and
a decoder pair. The encoder maps a variable-length source sequence to a
fixed-length vector, and the decoder maps the vector representation back
to a variable-length target sequence. The two networks are trained jointly
to maximize the conditional probability of the target sequence given a
source sequence. Additionally, we propose to use a rather sophisticated
hidden unit in order to improve both the memory capacity and the ease of
training.

The proposed RNN Encoder--Decoder with a novel hidden unit is empirically
evaluated on the task of translating from English to French. We train the model
to learn the translation probability of an English phrase to a corresponding
French phrase. The model is then used as a part of a standard phrase-based
SMT system by scoring each phrase pair in the phrase table.
The empirical evaluation reveals that this approach of scoring phrase pairs with
an RNN Encoder--Decoder improves the translation performance.

We qualitatively analyze the trained RNN Encoder--Decoder by comparing its
phrase scores with those given by the existing translation model.  The
qualitative analysis shows that the RNN Encoder--Decoder is better at capturing
the linguistic regularities in the phrase table, indirectly explaining the
quantitative improvements in the overall translation performance.  The further
analysis of the model reveals that the RNN Encoder--Decoder learns a 
continuous space representation of a phrase that preserves both the semantic
and syntactic structure of the phrase.

\section{RNN Encoder--Decoder}

\subsection{Preliminary: Recurrent Neural Networks}
\label{sec:rnn}

A recurrent neural network (RNN) is a neural network that consists of a hidden
state $\vh$ and an optional output $\vy$ which operates on a variable-length
sequence $\vx=(x_1, \dots, x_T)$. At each time step $t$, the hidden state
$\vh_{\qt{t}}$ of the RNN is updated by
\begin{align}
    \label{eq:encoding}
    \vh_{\qt{t}} = f\left( \vh_{\qt{t-1}}, x_t \right),
\end{align}
where $f$ is a non-linear activation function. $f$ may be as simple as an
element-wise logistic sigmoid function and as complex as a long short-term
memory (LSTM) unit~\mbox{\cite{Hochreiter1997}}.

An RNN can learn a probability distribution over a sequence by being trained to
predict the next symbol in a sequence. In that case, the output at each timestep
$t$ is the conditional distribution $p(x_t \mid x_{t-1}, \dots, x_1)$. For example,
a multinomial distribution ($1$-of-$K$ coding) can be output using a softmax
activation function

\begin{align}
    \label{eq:softmax}
    p(x_{t,j} = 1 \mid x_{t-1}, \dots, x_1) = \frac{\exp \left(
        \vw_j \vh_{\qt{t}}\right) } {\sum_{j'=1}^{K} \exp \left( \vw_{j'}
        \vh_{\qt{t}}\right) },
\end{align}
for all possible symbols $j=1,\dots,K$, where $\vw_j$ are the rows of a
weight matrix $\mW$. By combining these probabilities,
we can compute the probability of the sequence $\vx$ using
\begin{align}
    \label{eq:distribution}
    p(\vx) = \prod_{t=1}^T p(x_t \mid x_{t-1}, \dots, x_1).
\end{align}

From this learned distribution, it is straightforward to sample a new sequence
by iteratively sampling a symbol at each time step.

\subsection{RNN Encoder--Decoder}
\label{sec:encdec}

In this paper, we propose a novel neural network architecture that learns to
\textit{encode} a variable-length sequence into a fixed-length vector
representation and to \textit{decode} a given fixed-length vector representation
back into a variable-length sequence. From a probabilistic perspective, this
new model is a general method to learn the conditional distribution over a
variable-length sequence conditioned on yet another variable-length sequence,
e.g. $p(y_1, \dots, y_{T'} \mid x_1, \dots, x_T)$, where one should note that
the input and output sequence lengths $T$ and $T'$ may differ.

The encoder is an RNN that reads each symbol of an input sequence $\vx$
sequentially. As it reads each symbol, the hidden state of the RNN changes
according to Eq.~\eqref{eq:encoding}. After reading the end of the sequence
(marked by an end-of-sequence symbol), the hidden state of the RNN is a summary
$\vc$ of the whole input sequence.

The decoder of the proposed model is another RNN which is trained to
\textit{generate} the output sequence by predicting the next symbol $y_t$ given
the hidden state $\vh_{\qt{t}}$. However, unlike the RNN described in
Sec.~\mbox{\ref{sec:rnn}}, both $y_t$ and $\vh_{\qt{t}}$ are also conditioned on $y_{t-1}$
and on the summary $\vc$ of the input sequence.
Hence, the hidden state of the decoder at time
$t$ is computed by,
\begin{align*}
    \vh_{\qt{t}} = f\left( \vh_{\qt{t-1}}, y_{t-1}, \vc \right),
\end{align*}
and similarly, the conditional distribution of the next symbol is
\begin{align*}
    P(y_t | y_{t-1}, y_{t-2}, \ldots, y_1, \vc) = g\left(\vh_{\qt{t}}, y_{t-1}, \vc \right).
\end{align*}
for given activation functions $f$ and $g$ (the latter must produce valid probabilities,
e.g.\ with a softmax).

\begin{figure}
    \centering
    \includegraphics[width=0.8\columnwidth]{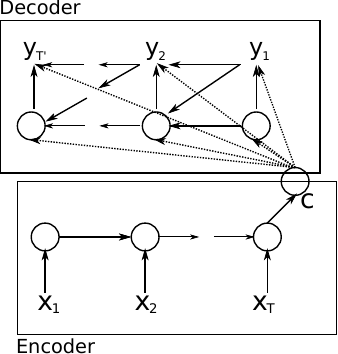}
    \caption{An illustration of the proposed RNN Encoder--Decoder.}
    \label{fig:encdec}
    \vspace{-3mm}
\end{figure}

See Fig.~\mbox{\ref{fig:encdec}} for a graphical depiction of the proposed
model architecture.

The two components of the proposed \textit{RNN Encoder--Decoder} are jointly
trained to maximize the conditional log-likelihood
\begin{align}
    \label{eq:loglik}
    \max_\TT \frac{1}{N} \sum_{n=1}^N \log p_{\TT}(\vy_n \mid \vx_n),
\end{align}
where $\TT$ is the set of the model parameters and each $\left(\vx_n,
\vy_n\right)$ is an (input sequence, output sequence) pair from the training set. 
In our case, as the
output of the decoder, starting from the input, is differentiable, we can use a
gradient-based algorithm to estimate the model parameters.

Once the RNN Encoder--Decoder is trained, the model can be used in two ways. One
way is to use the model to generate a target sequence given an input sequence.
On the other hand, the model can be used to \textit{score} a given pair of input
and output sequences, where the score is simply a probability $p_\TT(\vy \mid
\vx)$ from Eqs.~\eqref{eq:distribution} and \eqref{eq:loglik}. 

\subsection{Hidden Unit that Adaptively Remembers and Forgets}

\begin{figure}
    \centering
    \includegraphics[width=0.7\columnwidth]{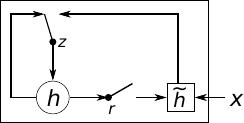}
    \caption{An illustration of the proposed hidden activation function. The
        update gate $z$ selects whether the hidden state is to be updated with a
        new hidden state $\tilde{h}$. The reset gate $r$ decides whether the
        previous hidden state is ignored. See
        Eqs.~\eqref{eq:reset}--\eqref{eq:preact} for the detailed equations of
        $r$, $z$, $h$ and $\tilde{h}$.}
    \label{fig:hidden}
    \vspace{-3mm}
\end{figure}

In addition to a novel model architecture, we also propose a new type of
hidden unit ($f$ in Eq.~\eqref{eq:encoding}) that has been motivated by the LSTM unit
but is much simpler to compute and implement.\footnote{
    The LSTM unit, which has shown impressive results in several applications
    such as speech recognition, has a memory cell and four gating units that
    adaptively control the information flow inside the unit, compared to only
    two gating units in the proposed hidden unit. For details on LSTM networks,
    see, e.g., \mbox{\cite{Graves-book2012}}.
}
Fig.~\mbox{\ref{fig:hidden}} shows the
graphical depiction of the proposed hidden unit.

Let us describe how the activation of the $j$-th hidden unit is computed. First, the
\textit{reset} gate $r_j$ is computed by
\begin{align}
    \label{eq:reset}
    r_j = \sigma\left( \left[ \mW_r \vx \right]_j + \left[\mU_r \vh_{\qt{t-1}}\right]_j \right),
\end{align}
where $\sigma$ is the logistic sigmoid function, and $\left[ . \right]_j$ denotes
the $j$-th element of a vector. $\vx$ and $\vh_{t-1}$ are the input and the
previous hidden state, respectively. $\mW_r$ and $\mU_r$ are weight matrices
which are learned.

Similarly, the \textit{update} gate $z_j$ is computed by
\begin{align}
    \label{eq:update}
    z_j = \sigma\left( \left[ \mW_z \vx \right]_j + \left[\mU_z \vh_{\qt{t-1}}\right]_j \right).
\end{align}

The actual activation of the proposed unit $h_j$ is then computed by
\begin{align}
    \label{eq:activation}
    h_j^{\qt{t}} = z_j h_j^{\qt{t-1}} + (1 - z_j) \tilde{h}_j^{\qt{t}},
\end{align}
where
\begin{align}
    \label{eq:preact}
    \tilde{h}_j^{\qt{t}} = \phi\left( 
    \left[ \mW \vx \right]_j + \left[ \mU \left( \vr \odot \vh_{\qt{t-1}} \right) \right]_j
    \right).
\end{align}

In this formulation, when the reset gate is close to 0, the hidden state is
forced to ignore the previous hidden state and reset with the current input
only. This effectively allows the hidden state to \textit{drop} any information
that is found to be irrelevant later in the future, thus, allowing a more
compact representation.  

On the other hand, the update gate controls how much information from the
previous hidden state will carry over to the current hidden state. This acts
similarly to the memory cell in the LSTM network and helps the RNN to remember long-term
information. Furthermore, this may be considered an adaptive variant of a
leaky-integration unit~\mbox{\cite{Bengio2013rec}}.

As each hidden unit has separate reset and update gates, each hidden unit will
learn to capture dependencies over different time scales. Those units that learn
to capture short-term dependencies will tend to have reset gates that are frequently
active, but those that capture longer-term dependencies will have update gates
that are mostly active.

In our preliminary experiments, we found that it is crucial to use this new unit
with gating units. We were not able to get meaningful result with an oft-used
$\tanh$ unit without any gating.

\section{Statistical Machine Translation}

In a commonly used statistical machine translation system (SMT), the goal of the
system (decoder, specifically) is to find a translation $\vf$ given a source
sentence $\ve$, which maximizes
\begin{align*}
    p(\vf \mid \ve) \propto  p(\ve \mid \vf) p(\vf),
\end{align*}
where the first term at the right hand side is called \textit{translation model}
and the latter \textit{language model} (see, e.g., \mbox{\cite{Koehn2005}}).
In practice, however, most SMT systems model $\log p(\vf \mid \ve)$ as a log-linear model with
additional features and corresponding weights:
\begin{align}
    \label{eq:loglinear}
    \log p(\vf \mid \ve) = \sum_{n=1}^N w_n f_n(\vf, \ve) + \log Z(\ve),
\end{align}
where $f_n$ and $w_n$ are the $n$-th feature and weight, respectively. $Z(\ve)$
is a normalization constant that does not depend on the weights. The weights are
often optimized to maximize the BLEU score on a development set.

In the phrase-based SMT framework introduced in \mbox{\cite{Koehn2003}} and
\mbox{\cite{Marcu2002}}, the translation model \mbox{$\log p(\ve\mid\vf)$} is factorized
into the translation probabilities of matching phrases in the source and target
sentences.\footnote{
    Without loss of generality, from here on, we refer to $p(\ve \mid \vf)$ for
    each phrase pair as a translation model as well
} These probabilities are once again considered additional features
in the log-linear model (see Eq.~\eqref{eq:loglinear}) and are weighted
accordingly to maximize the BLEU score.

Since the neural net language model was proposed in
\mbox{\cite{Bengio2003lm}}, neural networks have been used widely in SMT systems.
In many cases, neural networks have been used to \textit{rescore} translation
hypotheses ($n$-best lists) (see, e.g., \mbox{\cite{Schwenk2006}}). Recently,
however, there has been interest in training neural networks to score the
translated sentence (or phrase pairs) using a representation of the source
sentence as an additional input. See, e.g., \mbox{\cite{Schwenk2012}},
\mbox{\cite{Son2012}} and \mbox{\cite{Zou2013}}.

\subsection{Scoring Phrase Pairs with RNN Encoder--Decoder}
\label{sec:score_rnn}

Here we propose to train the RNN Encoder--Decoder (see
Sec.~\mbox{\ref{sec:encdec}}) on a table of phrase pairs and use its scores
as additional features in the log-linear model in Eq.~\eqref{eq:loglinear} when
tuning the SMT decoder.

When we train the RNN Encoder--Decoder, we ignore the (normalized) frequencies
of each phrase pair in the original corpora. This measure was taken in order (1)
to reduce the computational expense of randomly selecting phrase pairs from a
large phrase table according to the normalized frequencies and (2) to ensure
that the RNN Encoder--Decoder does not simply learn to rank the phrase pairs
according to their numbers of occurrences. One underlying reason for this choice
was that the existing translation probability in the phrase table already
reflects the frequencies of the phrase pairs in the original corpus. With a
fixed capacity of the RNN Encoder--Decoder, we try to ensure that most of
the capacity of the model is focused toward learning linguistic regularities,
i.e., distinguishing between plausible and implausible translations, or learning
the ``manifold'' (region of probability concentration) of plausible translations.

Once the RNN Encoder--Decoder is trained, we add a new score for each phrase
pair to the existing phrase table. This allows the new scores to enter
into the existing tuning algorithm with minimal additional overhead
in computation.

As Schwenk pointed out in \mbox{\cite{Schwenk2012}}, it is possible to
completely replace the existing phrase table with the proposed RNN
Encoder--Decoder. In that case, for a given source phrase, the RNN
Encoder--Decoder will need to generate a list of (good) target phrases. This
requires, however, an expensive sampling procedure to be performed repeatedly.
In this paper, thus, we only consider rescoring the phrase pairs in the phrase
table.

\subsection{Related Approaches: Neural Networks in Machine Translation}

Before presenting the empirical results, we discuss a number of recent works
that have proposed to use neural networks in the context of SMT.

Schwenk in \mbox{\cite{Schwenk2012}} proposed a similar approach of scoring
phrase pairs. Instead of the RNN-based neural network, he used a feedforward
neural network that has fixed-size inputs (7 words in his case, with
zero-padding for shorter phrases) and fixed-size outputs (7 words in the target
language). When it is used specifically for scoring phrases for the SMT system,
the maximum phrase length is often chosen to be small. However, as the length of
phrases increases or as we apply neural networks to other variable-length
sequence data, it is important that the neural network can handle
variable-length input and output. The proposed RNN Encoder--Decoder is
well-suited for these applications.

Similar to \mbox{\cite{Schwenk2012}}, Devlin et al. \mbox{\cite{Devlin2014}}
proposed to use a feedforward neural network to model a translation model,
however, by predicting one word in a target phrase at a time. They reported an
impressive improvement, but their approach still requires the maximum length of
the input phrase (or context words) to be fixed a priori.

Although it is not exactly a neural network they train, the authors of
\mbox{\cite{Zou2013}} proposed to learn a bilingual embedding of words/phrases.
They use the learned embedding to compute the distance between a pair of phrases
which is used as an additional score of the phrase pair in an SMT system.

In \mbox{\cite{Chandar2014}}, a feedforward neural network was trained to learn
a mapping from a bag-of-words representation of an input phrase to an output
phrase. This is closely related to both the proposed RNN Encoder--Decoder and
the model proposed in \mbox{\cite{Schwenk2012}}, except that their input
representation of a phrase is a bag-of-words. A similar approach of using
bag-of-words representations was proposed in \cite{Gao2013} as well.  Earlier,
a similar encoder--decoder model using two recursive neural networks was
proposed in \mbox{\cite{SocherEtAl2011:PoolRAE}}, but their model was
restricted to a monolingual setting, i.e.\ the model reconstructs an input
sentence. More recently, another encoder--decoder model using an RNN was
proposed in \cite{Auli2013}, where the decoder is conditioned on a
representation of either a source sentence or a source context.

One important difference between the proposed RNN Encoder--Decoder and the
approaches in \mbox{\cite{Zou2013}} and \mbox{\cite{Chandar2014}} is that the
order of the words in source and target phrases is taken into account. The RNN
Encoder--Decoder naturally distinguishes between sequences that have the same
words but in a different order, whereas the aforementioned approaches
effectively ignore order information.

The closest approach related to the proposed RNN Encoder--Decoder is the
Recurrent Continuous Translation Model (Model 2) proposed in
\mbox{\cite{Kalchbrenner2012}}. In their paper, they proposed a similar model
that consists of an encoder and decoder. The difference with our model is that
they used a convolutional $n$-gram model (CGM) for the encoder and the hybrid of
an inverse CGM and a recurrent neural network for the decoder. They, however,
evaluated their model on rescoring the $n$-best list proposed by the
conventional SMT system and computing the perplexity of the gold standard
translations.

\section{Experiments}

We evaluate our approach on the English/French translation task of the
WMT'14 workshop.

\subsection{Data and Baseline System}

Large amounts of resources are available to build an English/French SMT system
in the framework of the WMT'14 translation task.  The bilingual corpora include
Europarl (61M words), news commentary (5.5M), UN (421M), and two crawled
corpora of 90M and 780M words respectively.  The last two corpora are quite
noisy. To train the French language model, about 712M words of crawled
newspaper material is available in addition to the target side of the bitexts.
All the word counts refer to French words after tokenization.

It is commonly acknowledged that training statistical models on the concatenation
of all this data does not necessarily lead to optimal performance, and results 
in extremely large models which are difficult to handle.  Instead, one should
focus on the most relevant subset of the data for a given task.  We have done
so by applying the data selection method proposed in \mbox{\cite{Moore2010}}, and its extension to
bitexts \mbox{\cite{Axelrod2011}}. By these means we selected a subset of 418M words
out of more than 2G words for language modeling
and a subset of 348M out of 850M words for training the RNN Encoder--Decoder.  We
used the test set \texttt{newstest2012 and 2013} for data selection and weight
tuning with MERT, and \texttt{newstest2014} as our test set. Each set has more than 70
thousand words and a single reference translation.

For training the neural networks, including the proposed RNN Encoder--Decoder,
we limited the source and target vocabulary to the most frequent 15,000 words
for both English and French. This covers approximately 93\% of the dataset. All
the out-of-vocabulary words were mapped to a special token ($\left[ \text{UNK}
\right]$).

The baseline phrase-based SMT system was built using Moses with default
settings.  This system achieves a BLEU score of 30.64 and 33.3 on the
development and test sets, respectively (see Table~\mbox{\ref{tab:result}}).

\begin{table}
    \centering
    \begin{tabular}{l | c | c}
        \multirow{2}{*}{Models} & \multicolumn{2}{|c}{BLEU} \\
        & dev & test \\
        \hline\hline
        Baseline            & 30.64         & 33.30 \\
        RNN                 & 31.20         & 33.87 \\
        CSLM + RNN          & 31.48         & 34.64 \\
        CSLM + RNN + WP     & 31.50         & 34.54 \\
    \end{tabular}
    \caption{BLEU scores computed on the development and test sets using
        different combinations of approaches. WP denotes a \textit{word
        penalty}, where we penalizes the number of unknown words to neural
    networks.}
    \label{tab:result}
    \vspace{-3mm}
\end{table}

\subsubsection{RNN Encoder--Decoder}

The RNN Encoder--Decoder used in the experiment had 1000 hidden units with the
proposed gates at the encoder and at the decoder. The input matrix between
each input symbol $x_{\qt{t}}$ and the hidden unit is approximated with two
lower-rank matrices, and the output matrix is approximated similarly. We used
rank-100 matrices, equivalent to learning an embedding of dimension 100 for each word. 
The activation function used for $\tilde{h}$ in
Eq.~\eqref{eq:preact} is a hyperbolic tangent function.  The computation from
the hidden state in the decoder to the output is implemented as a deep neural
network~\mbox{\cite{Pascanu2014rec}} with a single intermediate layer having 500 maxout
units each pooling 2 inputs~\mbox{\cite{Goodfellow_maxout_2013}}. 

All the weight parameters in the RNN Encoder--Decoder were initialized by
sampling from an isotropic zero-mean (white) Gaussian distribution with its
standard deviation fixed to $0.01$, except for the recurrent weight parameters.
For the recurrent weight matrices, we first sampled from a white Gaussian
distribution and used its left singular vectors matrix,
following~\mbox{\cite{Saxe2014}}.

We used Adadelta and stochastic gradient descent to train the RNN
Encoder--Decoder with hyperparameters $\epsilon=10^{-6}$ and
$\rho=0.95$~\mbox{\cite{Zeiler-2012}}. At each update, we used 64 randomly
selected phrase pairs from a phrase table (which was created from 348M words).
The model was trained for approximately three days.

Details of the architecture used in the experiments are explained in more depth
in the supplementary material.

\begin{table*}[t]
    \centering
    \small
    \begin{tabular}{p{0.18\textwidth} | p{0.38\textwidth} | p{0.38\textwidth}}
        \hline
        Source & Translation Model & RNN Encoder--Decoder \\
        \hline\hline
        at the end of the & 
[a la fin de la] [ŕ la fin des années] [être supprimés à la fin de la] &
[à la fin du] [à la fin des] [à la fin de la]
        \\
        \hline
        for the first time &
        [{\bf r} © pour la premi{\bf r}\"ere fois] [été donnés pour la première fois] [été commémorée pour la première fois] & 
[pour la première fois] [pour la première fois ,] [pour la première fois que]  
        \\
        \hline
        in the United States and
        &
        [{\bf ?} aux  {\bf ?}tats-Unis et] [été ouvertes aux États-Unis et] [été constatées aux États-Unis et] &
[aux Etats-Unis et] [des Etats-Unis et] [des États-Unis et]
        \\
        \hline
        , as well as
        &
        [{\bf ?}s , qu'] [{\bf ?}s , ainsi que] [{\bf ?}re aussi bien que] & 
[, ainsi qu'] [, ainsi que] [, ainsi que les]  
        \\
        \hline
        one of the most
        &
[{\bf ?}t {\bf ?}l' un des plus] [{\bf ?}l' un des plus] [être retenue comme un de ses plus] &
[l' un des] [le] [un des]
        \\
        \hline
        \multicolumn{3}{c}{(a) Long, frequent source phrases}\\
        \multicolumn{3}{c}{~}\\
        \hline
        Source & Translation Model & RNN Encoder--Decoder \\
        \hline\hline
        , Minister of Communications and Transport &
[Secrétaire aux communications et aux transports :] [Secrétaire aux communications et aux transports] & 
[Secrétaire aux communications et aux transports] [Secrétaire aux communications et aux transports :]  
        \\
        \hline
        did not comply with the
        &
[vestimentaire , ne correspondaient pas à des] [susmentionnée n' était pas conforme aux] [présentées n' étaient pas conformes à la] & 
[n' ont pas respecté les] [n' était pas conforme aux] [n' ont pas respecté la]  
        \\
        \hline
        parts of the world .
        &
[© gions du monde .] [régions du monde considérées .] [région du monde considérée .] & 
[parties du monde .] [les parties du monde .] [des parties du monde .]  
        \\
        \hline
        the past few days .
        &
[le petit texte .] [cours des tout derniers jours .] [les tout derniers jours .] &
[ces derniers jours .] [les derniers jours .] [cours des derniers jours .]
        \\
        \hline
on Friday and Saturday
        &
[vendredi et samedi à la] [vendredi et samedi à] [se déroulera vendredi et samedi ,] &
[le vendredi et le samedi] [le vendredi et samedi] [vendredi et samedi]
        \\
        \hline
        \multicolumn{3}{c}{(b) Long, rare source phrases}\\
    \end{tabular}
    \caption{The top scoring target phrases for a small set of source phrases according to
    the translation model (direct translation probability) and by the RNN Encoder--Decoder. Source
    phrases were randomly selected from phrases with 4 or more words.
    {\bf ?} denotes an incomplete (partial) character. {\bf r} is a Cyrillic letter {\it ghe}.}
\label{tbl:samples}
    \vspace{-3mm}
\end{table*}

\subsubsection{Neural Language Model}

In order to assess the effectiveness of scoring phrase pairs with the proposed
RNN Encoder--Decoder, we also tried a more traditional approach of using a
neural network for learning a target language model
(CSLM)~\mbox{\cite{Schwenk2007}}.  Especially, the comparison between the SMT
system using CSLM and that using the proposed approach of phrase scoring by RNN
Encoder--Decoder will clarify whether the contributions from multiple neural
networks in different parts of the SMT system add up or are redundant.

We trained the CSLM model on 7-grams from the target corpus. Each input word
was projected into the embedding space $\RR^{512}$, and they were concatenated
to form a 3072-dimensional vector. The concatenated vector was fed through two
rectified layers (of size 1536 and 1024)~\mbox{\cite{Glorot+al-AI-2011-small}}.
The output layer was a simple softmax layer (see Eq.~\eqref{eq:softmax}). All
the weight parameters were initialized uniformly between $-0.01$ and $0.01$,
and the model was trained until the validation perplexity did not improve for
10 epochs. After training, the language model achieved a perplexity of 45.80.
The validation set was a random selection of 0.1\% of the corpus. The model was
used to score partial translations during the decoding process, which generally
leads to higher gains in BLEU score than n-best list
rescoring~\mbox{\cite{vaswanidecoding}}.

To address the computational complexity of using a CSLM in the decoder a buffer
was used to aggregate n-grams during the stack-search performed by the decoder.
Only when the buffer is full, or a stack is about to be pruned, the n-grams are
scored by the CSLM. This allows us to perform fast matrix-matrix multiplication
on GPU using Theano \mbox{\cite{bergstra+al:2010-scipy,Bastien-Theano-2012}}.

\begin{table*}[t]
    \centering
    \small
    \begin{tabular}{p{0.22\textwidth} | p{0.78\textwidth}}
        \hline
        Source & Samples from RNN Encoder--Decoder \\
        \hline\hline
        at the end of the & 
        [à la fin de la] ($\times 11$)
        \\
        \hline
        for the first time &
        [pour la première fois] ($\times 24$) [pour la première fois que] ($\times 2$)
        \\
        \hline
        in the United States and
        &
        [aux États-Unis et] ($\times 6$) [dans les États-Unis et] ($\times 4$)
        \\
        \hline
        , as well as
        &
        [, ainsi que] [,] [ainsi que] [, ainsi qu'] [et UNK]
        \\
        \hline
        one of the most
        &
        [l' un des plus] ($\times 9$) [l' un des] ($\times 5$) [l' une des plus] ($\times 2$)
        \\
        \hline
        \multicolumn{2}{c}{(a) Long, frequent source phrases}\\
        \multicolumn{2}{c}{~}\\
        \hline
        Source & Samples from RNN Encoder--Decoder \\
        \hline\hline
        , Minister of Communications and Transport &
[ , ministre des communications et le transport] ($\times 13$) 
        \\
        \hline
        did not comply with the
        &
        [n' tait pas conforme aux] [n' a pas respect l'] ($\times 2$) [n' a pas respect la] ($\times 3$)
        \\
        \hline
        parts of the world .
        &
        [arts du monde .] ($\times 11$) [des arts du monde .] ($\times 7$)
        \\
        \hline
        the past few days .
        &
        [quelques jours .] ($\times 5$) [les derniers jours .] ($\times 5$) [ces derniers jours .] ($\times 2$)
        \\
        \hline
on Friday and Saturday
        &
        [vendredi et samedi] ($\times 5$) [le vendredi et samedi] ($\times 7$) [le vendredi et le samedi] ($\times 4$)
        \\
        \hline
        \multicolumn{2}{c}{(b) Long, rare source phrases}\\
    \end{tabular}
    \caption{Samples generated from the RNN Encoder--Decoder for each source
        phrase used in Table~\mbox{\ref{tbl:samples}}. We show the top-5 target phrases
    out of 50 samples.  They are sorted by the RNN Encoder--Decoder scores.}
\label{tbl:samples_generated}
    \vspace{-3mm}
\end{table*}

\begin{figure}[h]
    \centering
    \includegraphics[width=0.9\columnwidth]{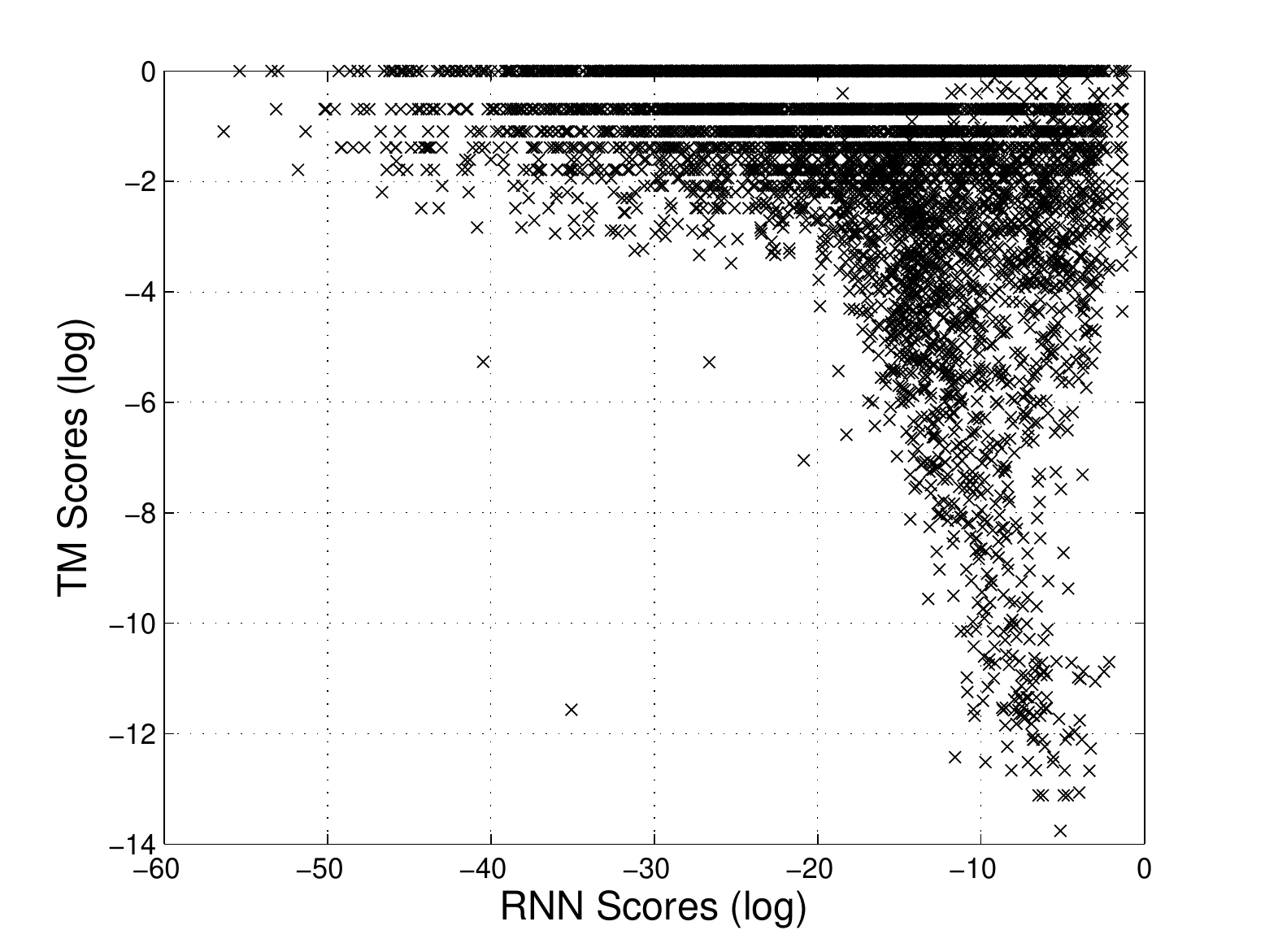}
    \caption{
        The visualization of phrase pairs according to their scores
        (log-probabilities) by the RNN Encoder--Decoder and the translation
        model. 
    }
    \label{fig:scores}
    \vspace{-3mm}
\end{figure}

\subsection{Quantitative Analysis}

We tried the following combinations:
\begin{enumerate}
    \itemsep -0.7em
    \item Baseline configuration
    \item Baseline + RNN
    \item Baseline + CSLM + RNN
    \item Baseline + CSLM + RNN + Word penalty
\end{enumerate}

The results are presented in Table~\mbox{\ref{tab:result}}. As expected, adding
features computed by neural networks consistently improves the performance over
the baseline performance. 

The best performance was achieved when we used both CSLM and the phrase scores
from the RNN Encoder--Decoder. This suggests that the contributions of the CSLM
and the RNN Encoder--Decoder are not too correlated and that one can expect
better results by improving each method independently.  Furthermore, we tried
penalizing the number of words that are unknown to the neural networks (i.e.\
words which are not in the shortlist). We do so by simply adding the number of
unknown words as an additional feature the log-linear model in
Eq.~\eqref{eq:loglinear}.\footnote{
    To understand the effect of the penalty, consider the set of all words in the
    15,000 large shortlist, $\text{SL}$. All words $x^i \notin \text{SL}$ are
    replaced by a special token $\left[\text{UNK}\right]$ before being scored by
    the neural networks. Hence, the conditional probability of any $x_t^i \notin
    \text{SL}$ is actually given by the model as
    \begin{align*}
        p\left(x_t =\right.&\left. \left[\text{UNK}\right] \mid x_{<t}\right) = p\left(x_t \notin \text{SL} \mid x_{<t}\right) \\
    &= \sum_{x^j_t \notin SL} p\left(x_t^j \mid x_{<t} \right) \geq p\left(x_t^i \mid x_{<t} \right),
    \end{align*}
    where $x_{<t}$ is a shorthand notation for $x_{t-1},\dots,x_1$.

    As a result, the probability of words not in the shortlist is always
    overestimated. It is possible to address this issue by backing off to an
    existing model that contain non-shortlisted words (see
    \mbox{\cite{Schwenk2007}}) In this paper, however, we opt for introducing a
    word penalty instead, which counteracts the word probability overestimation.
}
However, in this case we were not able to achieve better performance on the test
set, but only on the development set.

\begin{figure*}[ht]
    \centering
    \begin{minipage}{0.48\textwidth}
        \centering
        \includegraphics[width=0.99\textwidth]{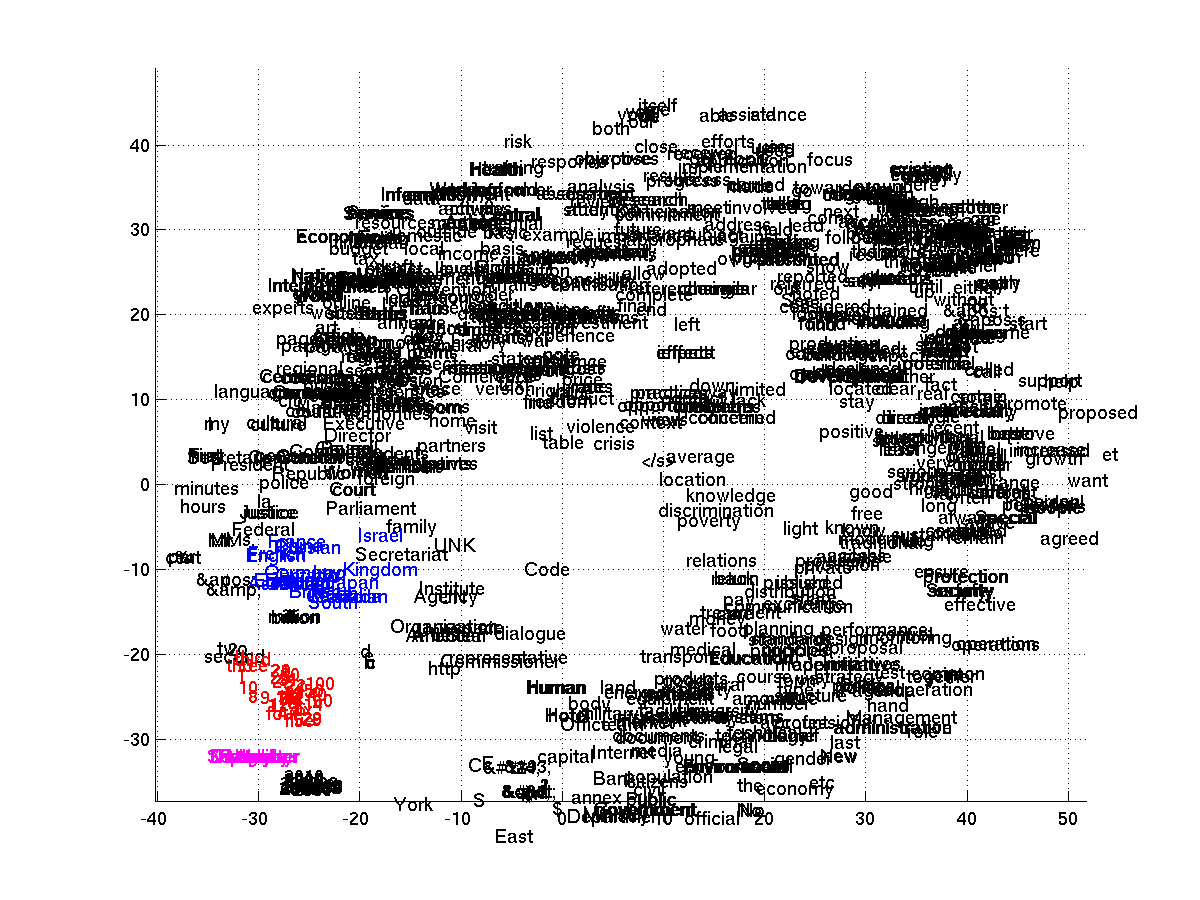}
    \end{minipage}
    \hfill
    \begin{minipage}{0.48\textwidth}
        \centering
        \includegraphics[width=0.99\textwidth]{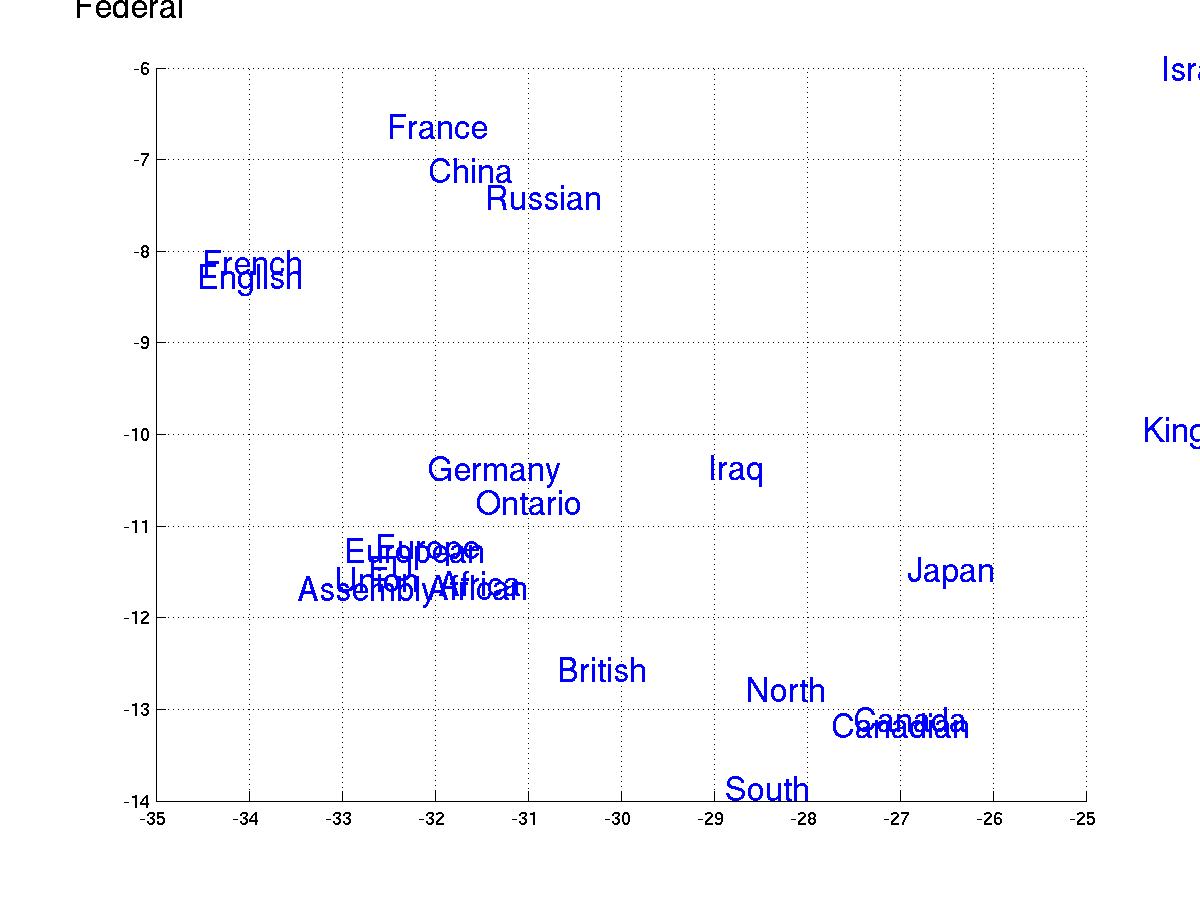}
    \end{minipage}
    \caption{2--D embedding of the learned word representation. The left
    one shows the full embedding space, while the right one shows a
    zoomed-in view of one region (color--coded). For more plots, see the
supplementary material.} 
    \label{fig:word_embed}
    \vspace{-3mm}
\end{figure*}

\begin{figure*}[ht]
    \centering
    \begin{minipage}{0.48\textwidth}
        \centering
        \includegraphics[width=0.99\textwidth]{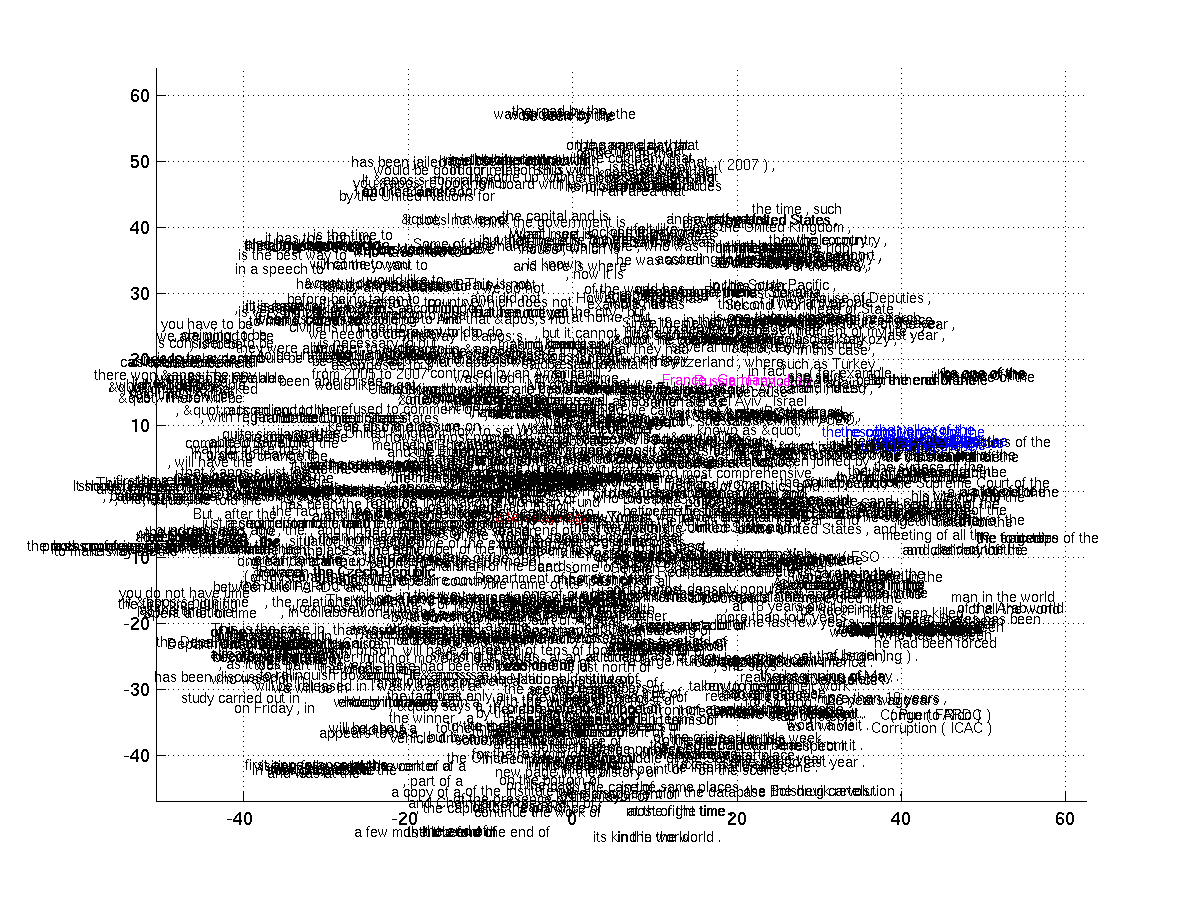}
    \end{minipage}
    \hfill
    \begin{minipage}{0.48\textwidth}
        \centering
        \includegraphics[width=0.99\textwidth]{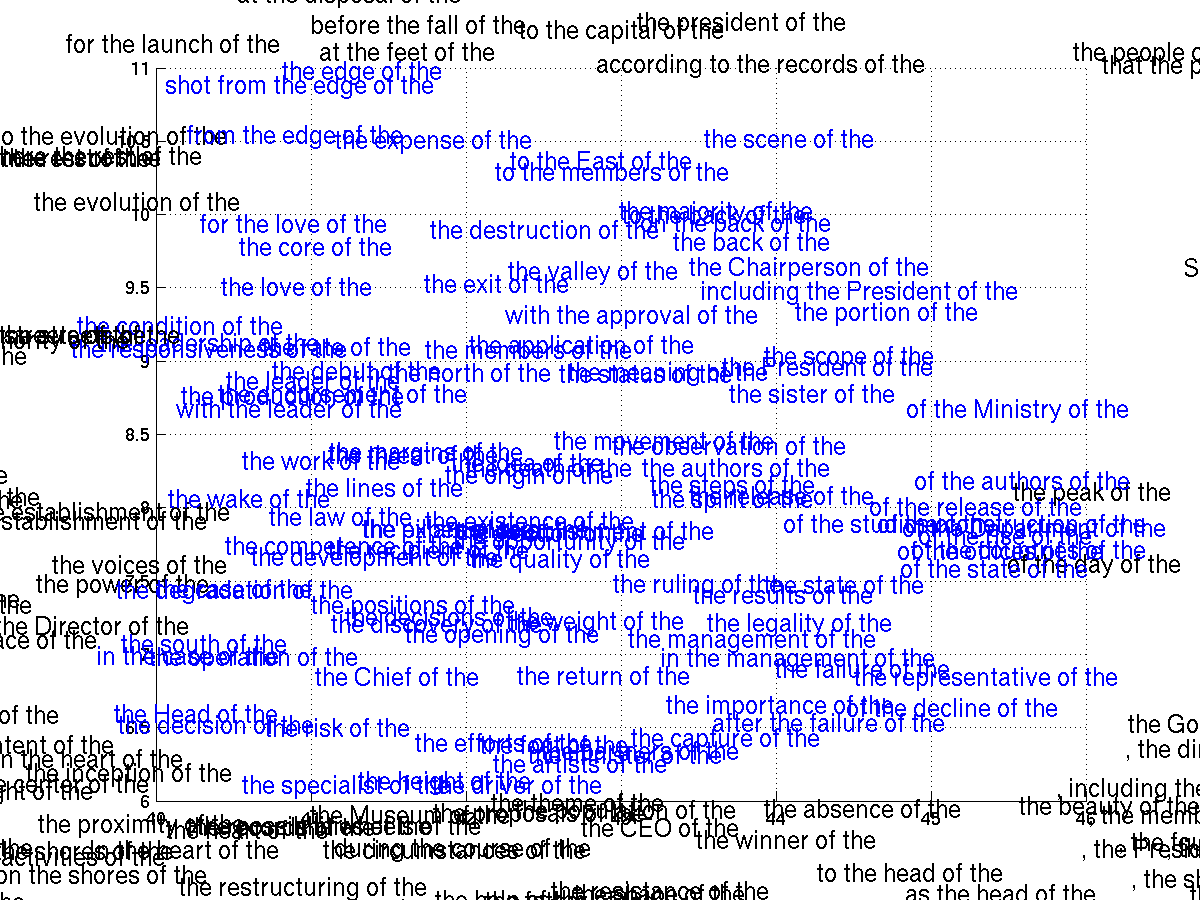}
    \end{minipage}
    \\
    \begin{minipage}{0.48\textwidth}
        \centering
        \includegraphics[width=0.99\textwidth]{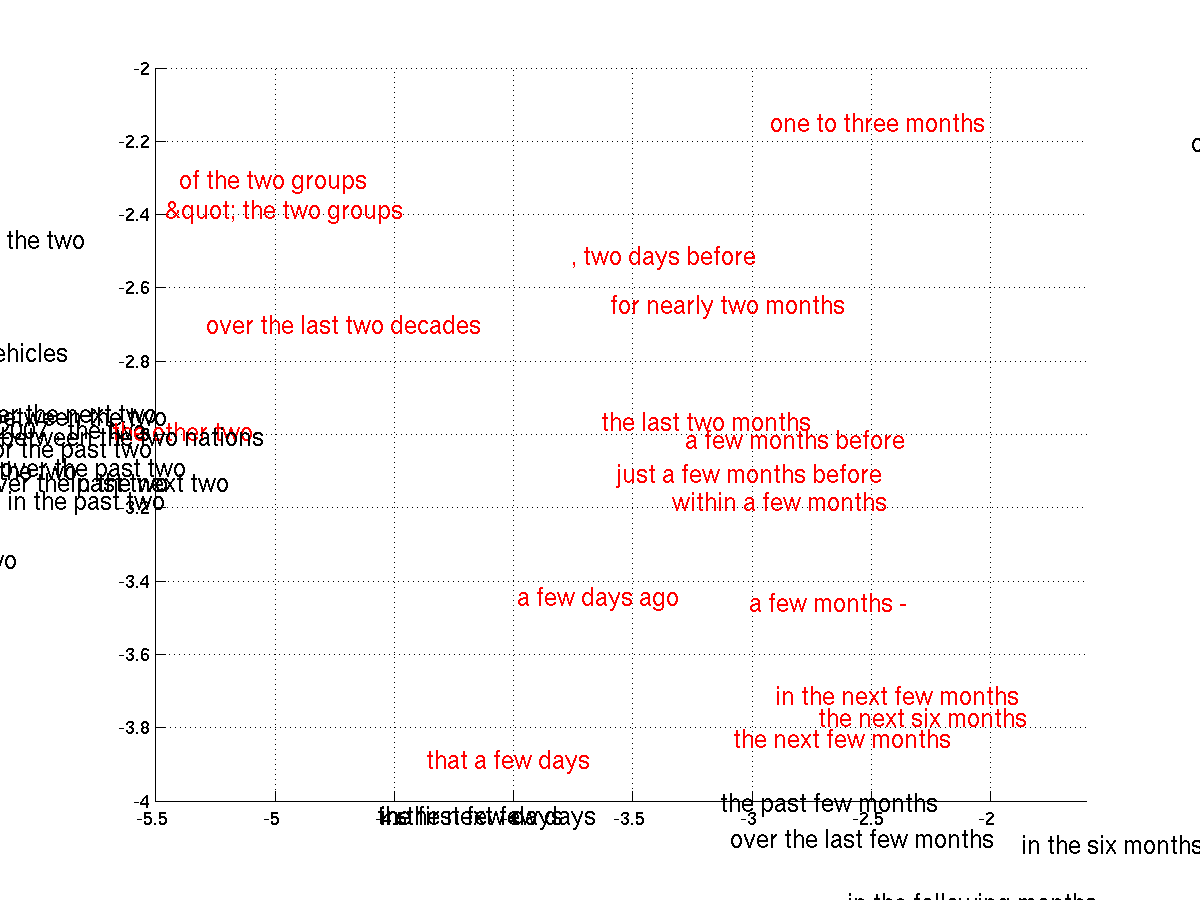}
    \end{minipage}
    \hfill
    \begin{minipage}{0.48\textwidth}
        \centering
        \includegraphics[width=0.99\textwidth]{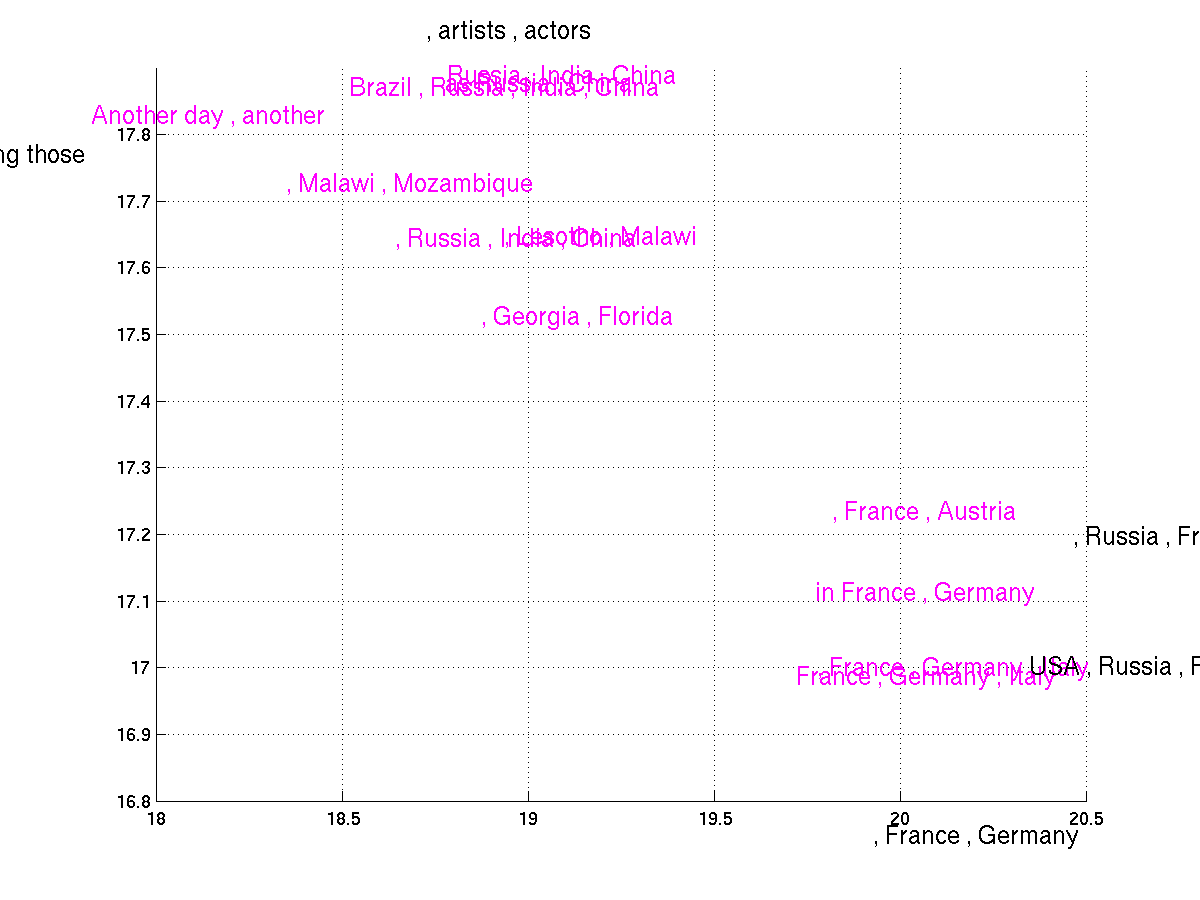}
    \end{minipage}
    \caption{2--D embedding of the learned phrase representation. The top left
    one shows the full representation space (5000 randomly selected points),
while the other three figures show the zoomed-in view of specific regions
(color--coded).} 
    \label{fig:phrase_embed}
    \vspace{-3mm}
\end{figure*}

\subsection{Qualitative Analysis}

In order to understand where the performance improvement comes from, we analyze
the phrase pair scores computed by the RNN Encoder--Decoder against the
corresponding
$p(\vf \mid \ve)$ from the translation model.  Since the existing translation model
relies solely on the statistics of the phrase pairs in the corpus, we expect
its scores to be better estimated for the frequent phrases but badly estimated
for rare phrases.  Also, as we mentioned earlier in
Sec.~\mbox{\ref{sec:score_rnn}}, we further expect the RNN Encoder--Decoder
which was trained without any frequency information to score the phrase pairs
based rather on the linguistic regularities than on the statistics of their
occurrences in the corpus.

We focus on those pairs whose source phrase is \textit{long} (more than 3 words
per source phrase) and \textit{frequent}. For each such source phrase, we look
at the target phrases that have been scored high either by the translation
probability $p(\vf \mid \ve)$ or by the RNN Encoder--Decoder.  Similarly, we
perform the same procedure with those pairs whose source phrase is \textit{long}
but \textit{rare} in the corpus.

Table~\mbox{\ref{tbl:samples}} lists the top-$3$ target phrases per source
phrase favored either by the translation model or by the RNN Encoder--Decoder.
The source phrases were randomly chosen among long ones having more than 4 or 5
words.

In most cases, the choices of the target phrases by the RNN Encoder--Decoder
are closer to actual or literal translations. We can observe that the RNN
Encoder--Decoder prefers shorter phrases in general. 

Interestingly, many phrase pairs were scored similarly by both the translation
model and the RNN Encoder--Decoder, but there were as many other phrase pairs
that were scored radically different (see Fig.~\mbox{\ref{fig:scores}}). This could 
arise from the proposed approach of training the RNN Encoder--Decoder on a
set of unique phrase pairs, discouraging the RNN Encoder--Decoder from learning
simply the frequencies of the phrase pairs from the corpus, as 
explained earlier.

Furthermore, in Table~\mbox{\ref{tbl:samples_generated}}, we show for each of
the source phrases in Table~\mbox{\ref{tbl:samples}}, the generated samples
from the RNN Encoder--Decoder. For each source phrase, we generated 50 samples
and show the top-five phrases accordingly to their scores. We can see that the
RNN Encoder--Decoder is able to propose well-formed target phrases without
looking at the actual phrase table. Importantly, the generated phrases do not
overlap completely with the target phrases from the phrase table.  This
encourages us to further investigate the possibility of replacing the whole or
a part of the phrase table with the proposed RNN Encoder--Decoder in the
future.

\subsection{Word and Phrase Representations}
\label{sec:rep}

Since the proposed RNN Encoder--Decoder is not specifically designed only for
the task of machine translation, here we briefly look at the properties of the
trained model. 

It has been known for some time that continuous space language models using
neural networks are able to learn semantically meaningful embeddings (See, e.g.,
\mbox{\cite{Bengio2003lm,Mikolov2013}}).  Since the proposed RNN Encoder--Decoder also
projects to and maps back from a sequence of words into a continuous space
vector, we expect to see a similar property with the proposed model as well.

The left plot in Fig.~\mbox{\ref{fig:word_embed}} shows the 2--D embedding of the
words using the word embedding matrix learned by the RNN Encoder--Decoder. The
projection was done by the recently proposed Barnes-Hut-SNE~\mbox{\cite{Maaten2013}}.
We can clearly see that semantically similar words are clustered with each other
(see the zoomed-in plots in Fig.~\mbox{\ref{fig:word_embed}}). 

The proposed RNN Encoder--Decoder naturally generates a continuous-space
representation of a phrase. The representation ($\vc$ in Fig.~\mbox{\ref{fig:encdec}})
in this case is a 1000-dimensional vector. Similarly to the word
representations, we visualize the representations of the phrases that consists of
four or more words using the Barnes-Hut-SNE in Fig.~\mbox{\ref{fig:phrase_embed}}.

From the visualization, it is clear that the RNN Encoder--Decoder captures
\textit{both semantic and syntactic} structures of the phrases. For instance,
in the bottom-left plot, most of the phrases are about the duration of time,
while those phrases that are syntactically similar are clustered together. 
The bottom-right plot shows the cluster of phrases that are semantically similar
(countries or regions). On the other hand, the top-right plot shows the phrases
that are syntactically similar.

\section{Conclusion}

In this paper, we proposed a new neural network architecture, called an
\textit{RNN Encoder--Decoder} that is able to learn the mapping from a sequence
of an arbitrary length to another sequence, possibly from a different set, of an
arbitrary length. The proposed RNN Encoder--Decoder is able to either score a
pair of sequences (in terms of a conditional probability) or generate a target
sequence given a source sequence.
Along with the new architecture, we proposed a novel hidden unit that includes a
reset gate and an update gate that adaptively control how much each hidden unit
remembers or forgets while reading/generating a sequence. 

We evaluated the proposed model with the task of statistical machine
translation, where we used the RNN Encoder--Decoder to score each phrase pair in
the phrase table. Qualitatively, we were able to show that the new model is
able to capture linguistic regularities in the phrase pairs well and also that
the RNN Encoder--Decoder is able to propose well-formed target phrases. 

The scores by the RNN Encoder--Decoder were found to improve the overall
translation performance in terms of BLEU scores. Also, we found that the
contribution by the RNN Encoder--Decoder is rather orthogonal to the existing
approach of using neural networks in the SMT system, so that we can improve
further the performance by using, for instance, the RNN Encoder--Decoder and the
neural net language model together.

Our qualitative analysis of the trained model shows that it indeed captures the
linguistic regularities in multiple levels i.e. at the word level as well as phrase level.
This suggests that there may be more natural language related applications that
may benefit from the proposed RNN Encoder--Decoder.

The proposed architecture has large potential for further improvement and
analysis. One approach that was not investigated here is to replace the whole,
or a part of the phrase table by letting the RNN Encoder--Decoder propose
target phrases. Also, noting that the proposed model is not limited to being
used with written language, it will be an important future research to apply the
proposed architecture to other applications such as speech transcription.

\section*{Acknowledgments}

KC, BM, CG, DB and YB would like to thank  NSERC, Calcul Qu\'{e}bec, Compute
Canada, the Canada Research Chairs and CIFAR. FB and HS were partially funded
by the European Commission under the project MateCat, and by DARPA under the
BOLT project.

\bibliographystyle{acl}
\bibliography{strings,strings-shorter,ml,aigaion,myref}

\newpage
\appendix
{
\onecolumn
\section{RNN Encoder--Decoder}
\label{sec:detail}

In this document, we describe in detail the architecture of the RNN
Encoder--Decoder used in the experiments.

Let us denote an source phrase by $X=\left( \vx_1, \vx_2, \dots, \vx_N \right)$
and a target phrase by $Y=\left( \vy_1, \vy_2, \dots, \vy_M \right)$. Each
phrase is a sequence of $K$-dimensional one-hot vectors, such that only one
element of the vector is $1$ and all the others are $0$. The index of the active
($1$) element indicates the word represented by the vector.

\subsection{Encoder}

Each word of the source phrase is embedded in a $500$-dimensional vector space:
$e(\vx_i) \in \RR^{500}$. $e(\vx)$ is used in Sec.~\ref{sec:rep} to visualize
the words.

The hidden state of an encoder consists of $1000$ hidden units,
and each one of them at time $t$ is computed by
\begin{align*}
    h_j^{\qt{t}} = z_j h_j^{\qt{t-1}} + (1 - z_j) \tilde{h}_j^{\qt{t}},
\end{align*}
where
\begin{align*}
    \tilde{h}_j^{\qt{t}} =& \tanh\left( \left[ \mW e(\vx_t) \right]_j + \left[
    \mU \left( \vr \odot \vh_{\qt{t-1}}\right) \right]_j \right),
    \\
    z_j =& \sigma\left( \left[ \mW_z e(\vx_t) \right]_j + 
    \left[\mU_z \vh_{\qt{t-1}}\right]_j \right),
    \\
    r_j =& \sigma\left( \left[ \mW_r e(\vx_t) \right]_j + 
    \left[\mU_r \vh_{\qt{t-1}}\right]_j \right).
\end{align*}
$\sigma$ and $\odot$ are a logistic sigmoid function and an element-wise
multiplication, respectively. To make the equations uncluttered, we omit
biases. The initial hidden state $h_j^{\qt{0}}$ is fixed to $0$.

Once the hidden state at the $N$ step (the end of the source
phrase) is computed, the representation of the source phrase
$\vc$ is
\begin{align*}
    \vc = \tanh \left( \mV \vh^{\qt{N}} \right).
\end{align*}

\subsubsection{Decoder}

The decoder starts by initializing the hidden state with
\begin{align*}
    {\vh'}^{\qt{0}} = \tanh \left( \mV' \vc \right),
\end{align*}
where we will use $\cdot'$ to distinguish parameters of the
decoder from those of the encoder.

The hidden state at time $t$ of the decoder is computed by
\begin{align*}
    {h'}_j^{\qt{t}} = {z'}_j {h'}_j^{\qt{t-1}} +
    (1 - {z'}_j) \tilde{h'}_j^{\qt{t}},
\end{align*}
where
\begin{align*}
    \tilde{h'}_j^{\qt{t}} =& \tanh\left( 
    \left[ \mW' e(\vy_{t-1}) \right]_j + 
    {r'}_j \left[ \mU' {\vh'}_{\qt{t-1}} +
    \mC \vc 
    \right]
    \right),
    \\
    {z'}_j =& \sigma\left( \left[ {\mW'}_z e(\vy_{t-1}) \right]_j + 
    \left[{\mU'}_z {\vh'}_{\qt{t-1}}\right]_j +
    \left[\mC_z \vc \right]_j 
    \right),
    \\
    {r'}_j =& \sigma\left( \left[ {\mW'}_r e(\vy_{t-1}) \right]_j + 
    \left[{\mU'}_r {\vh'}_{\qt{t-1}}\right]_j +
    \left[ \mC_r \vc \right]_j
    \right),
\end{align*}
and $e(\vy_{0})$ is an all-zero vector. Similarly to the case of
the encoder, $e(\vy)$ is an embedding of a target word.

Unlike the encoder which simply encodes the source phrase, the
decoder is learned to generate a target phrase. At each time $t$,
the decoder computes the probability of generating $j$-th word by
\begin{align*}
    p(y_{t,j} = 1 \mid \vy_{t-1}, \dots, \vy_1, X) = \frac{\exp
        \left( {\vg}_j \vs_{\qt{t}}\right) } {\sum_{j'=1}^{K}
        \exp \left( \vg_{j'} \vs_{\qt{t}}\right) },
\end{align*}
where the $i$-element of $\vs_{\qt{t}}$ is 
\begin{align*}
    s_i^{\qt{t}} = \max\left\{
    {s'}_{2i-1}^{\qt{t}}, {s'}_{2i}^{\qt{t}}
\right\}
\end{align*}
and
\begin{align*}
    {\vs'}^{\qt{t}} = \mO_h {\vh'}^{\qt{t}} + \mO_y \vy_{t-1} + \mO_c \vc.
\end{align*}
In short, the $s_i^{\qt{t}}$ is a so-called \textit{maxout} unit.

For the computational efficiency, instead of a single-matrix
output weight $\mG$, we use a product of two matrices such that
\begin{align*}
    \mG = \mG_l \mG_r,
\end{align*}
where $\mG_l \in \RR^{K \times 500}$ and $\mG_r \in \RR^{500 \times
1000}$.

\section{Word and Phrase Representations}
\label{sec:word_phrase_embed}

Here, we show enlarged plots of the word and phrase
representations in
Figs.~\ref{fig:word_embed}--\ref{fig:phrase_embed}.

\newpage
\begin{landscape}
\begin{figure*}[p]
    \iftoggle{arxiv}{
        \vspace{-20mm}
    }
    \centering
    \begin{minipage}{0.75\textwidth}
        \centering
        \includegraphics[width=0.99\textwidth]{word_all.png}
    \end{minipage}
    \begin{minipage}{0.75\textwidth}
        \centering
        \includegraphics[width=0.99\textwidth]{word_countries.png}
    \end{minipage}
    \\
    \begin{minipage}{0.75\textwidth}
        \centering
        \includegraphics[width=0.99\textwidth]{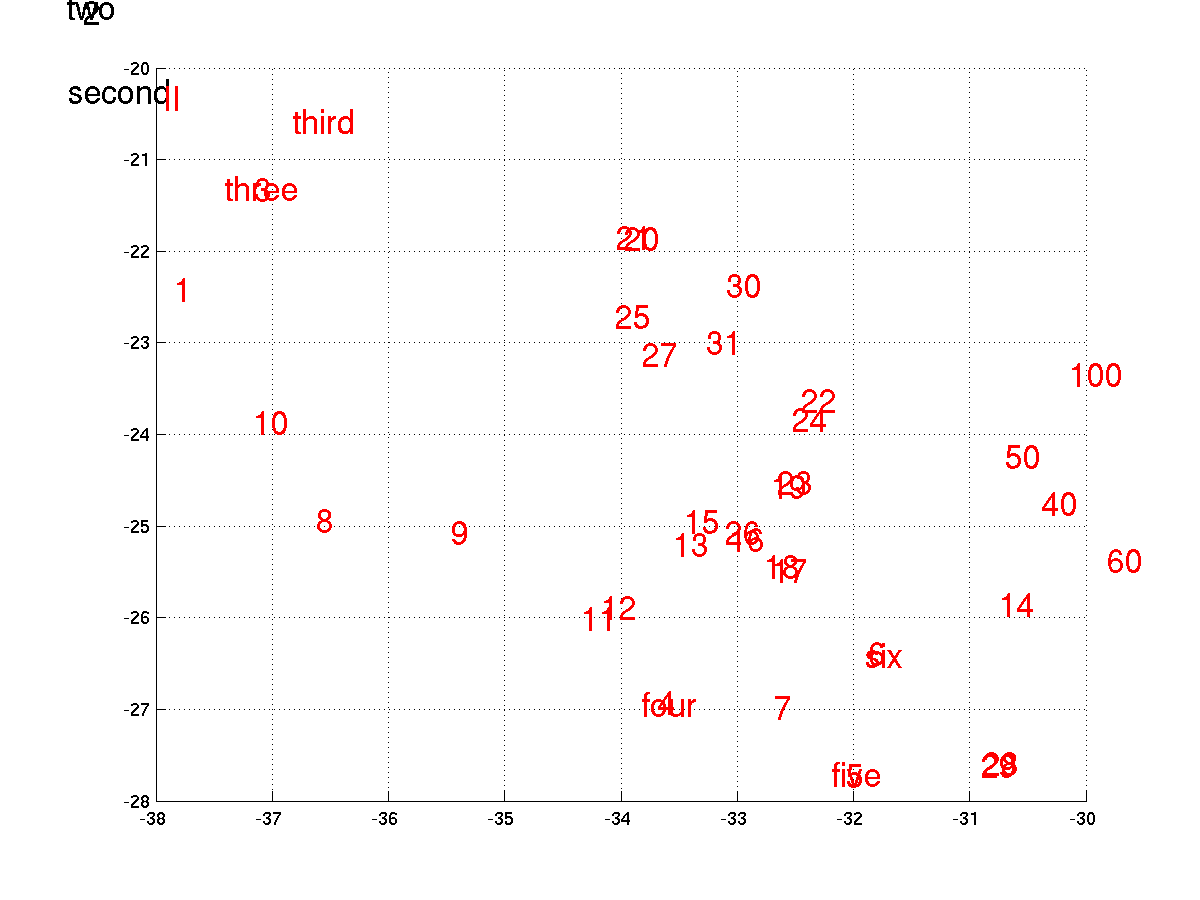}
    \end{minipage}
    \begin{minipage}{0.75\textwidth}
        \centering
        \includegraphics[width=0.99\textwidth]{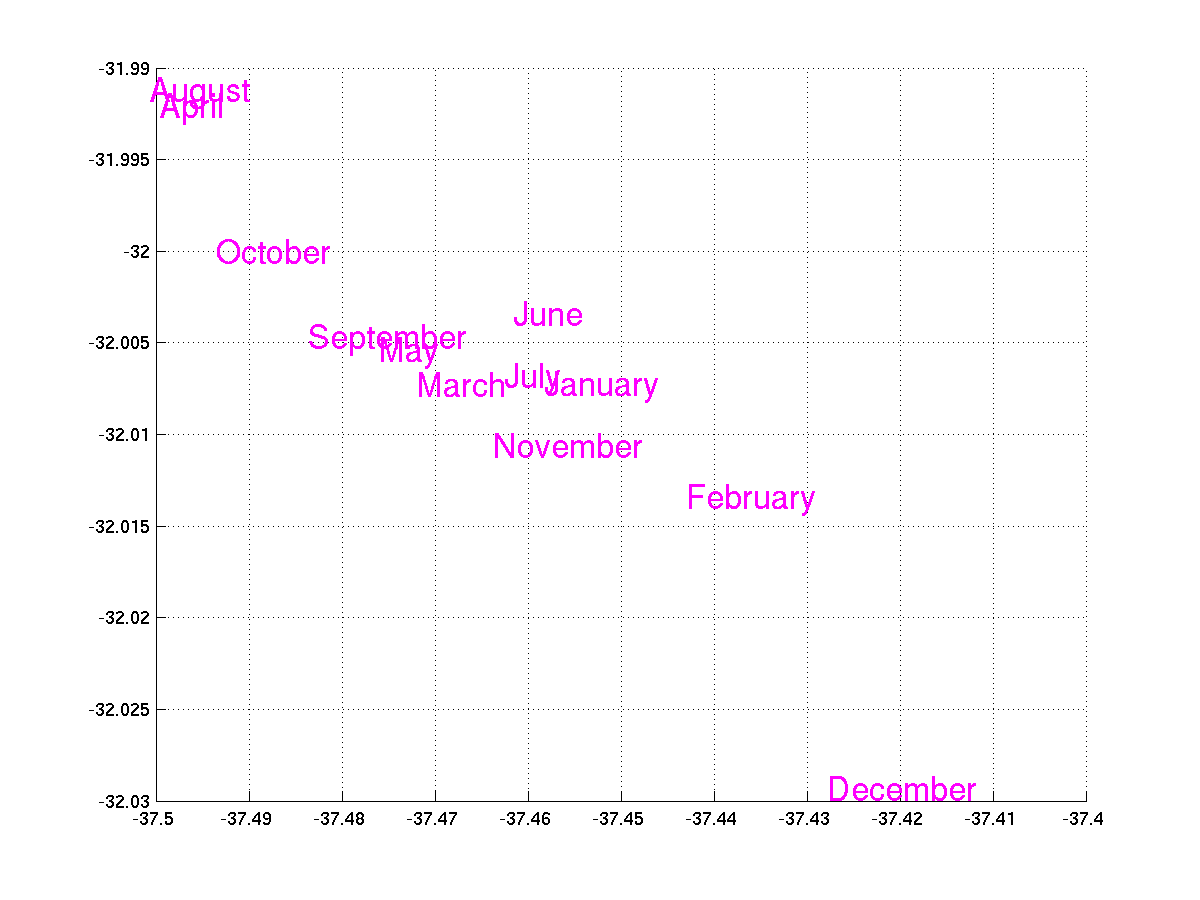}
    \end{minipage}
    \caption{2--D embedding of the learned word representation. The top left one
        shows the full embedding space, while the other three figures show the
    zoomed-in view of specific regions (color--coded).} 
\end{figure*}
\end{landscape}

\newpage
\begin{landscape}
\begin{figure*}[p]
    \iftoggle{arxiv}{
        \vspace{-20mm}
    }
    \centering
    \begin{minipage}{0.75\textwidth}
        \centering
        \includegraphics[width=0.99\textwidth]{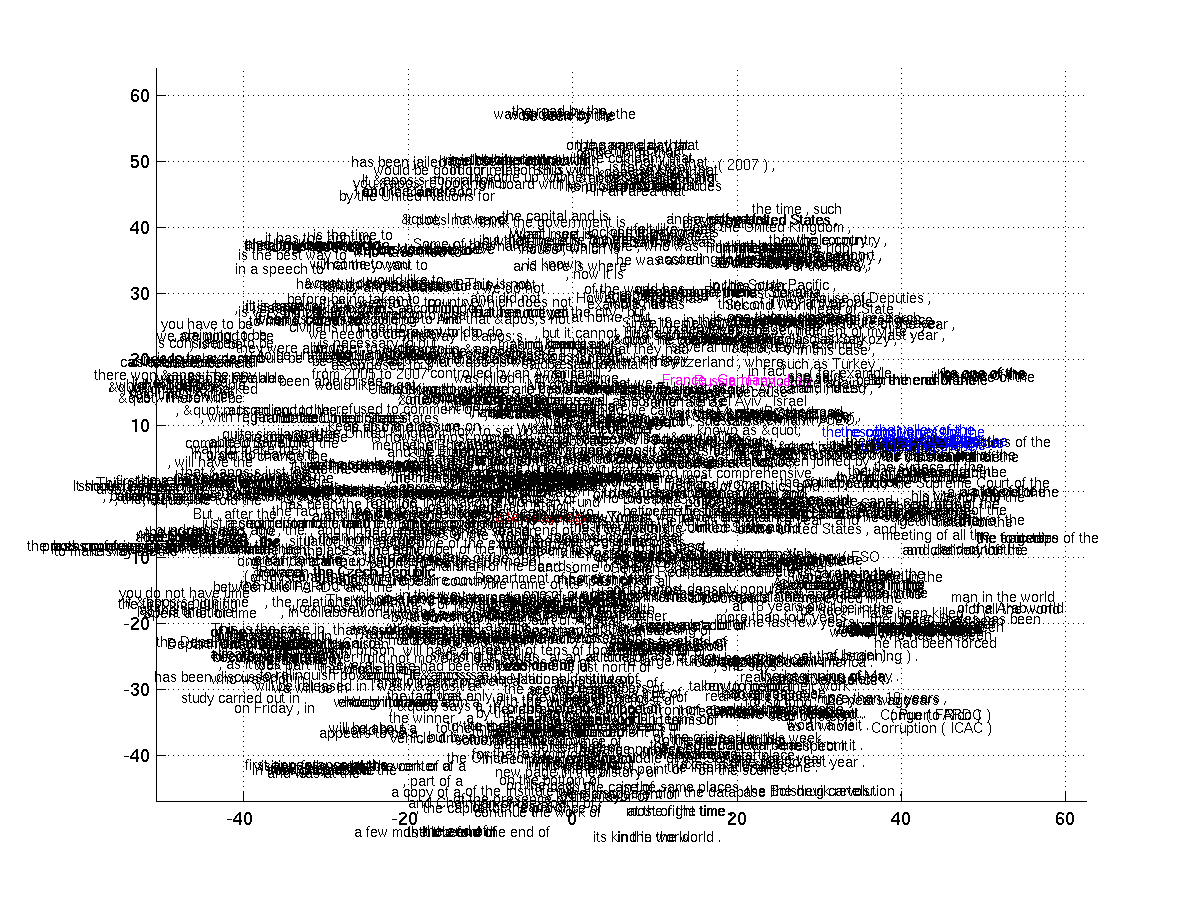}
    \end{minipage}
    \begin{minipage}{0.75\textwidth}
        \centering
        \includegraphics[width=0.99\textwidth]{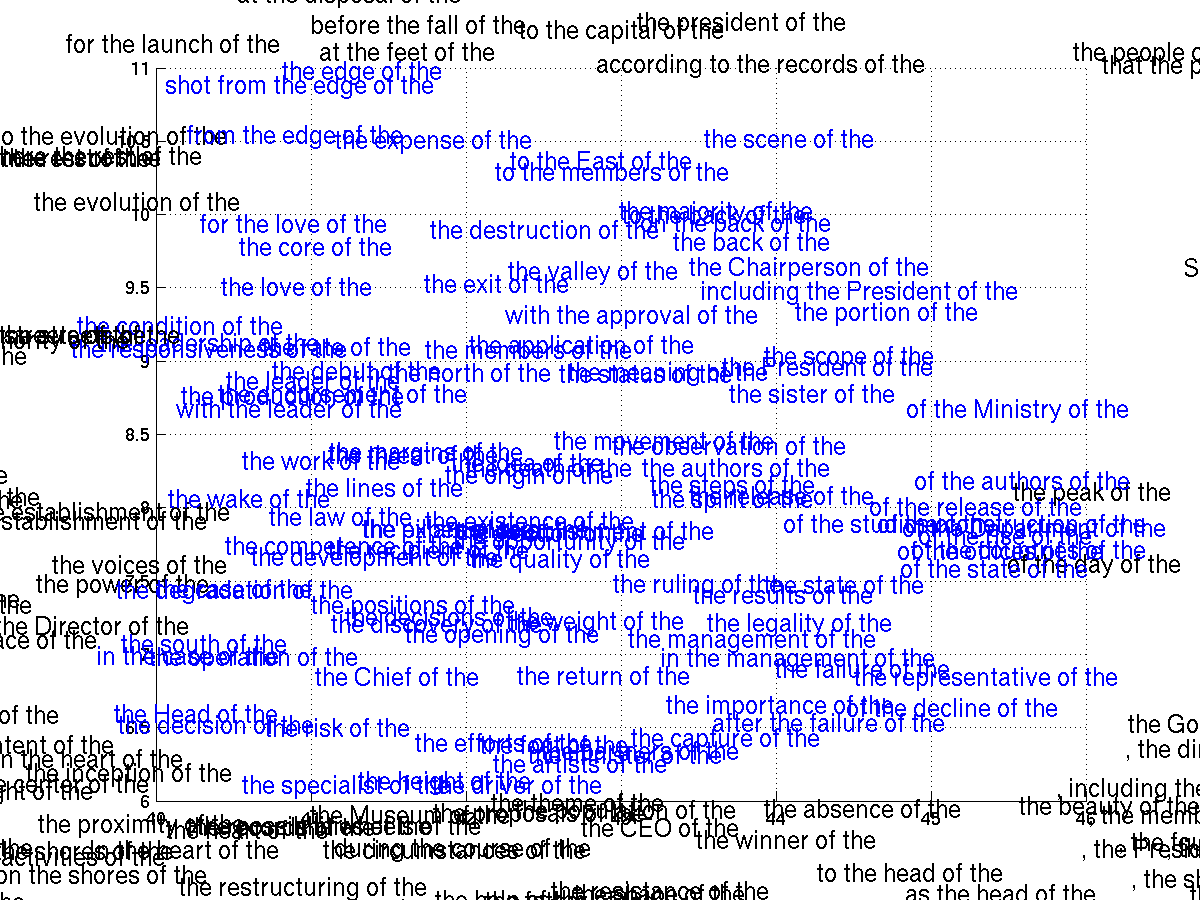}
    \end{minipage}
    \\
    \begin{minipage}{0.75\textwidth}
        \centering
        \includegraphics[width=0.99\textwidth]{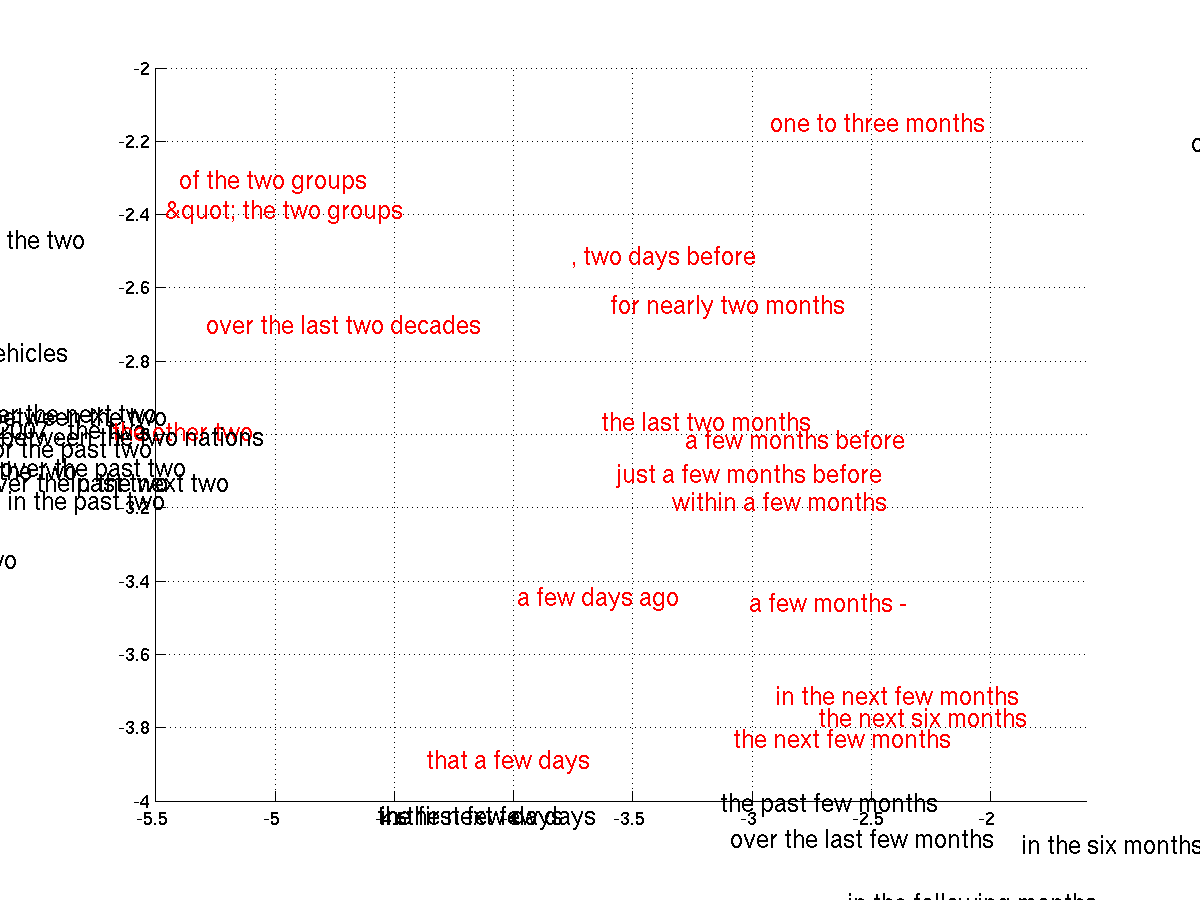}
    \end{minipage}
    \begin{minipage}{0.75\textwidth}
        \centering
        \includegraphics[width=0.99\textwidth]{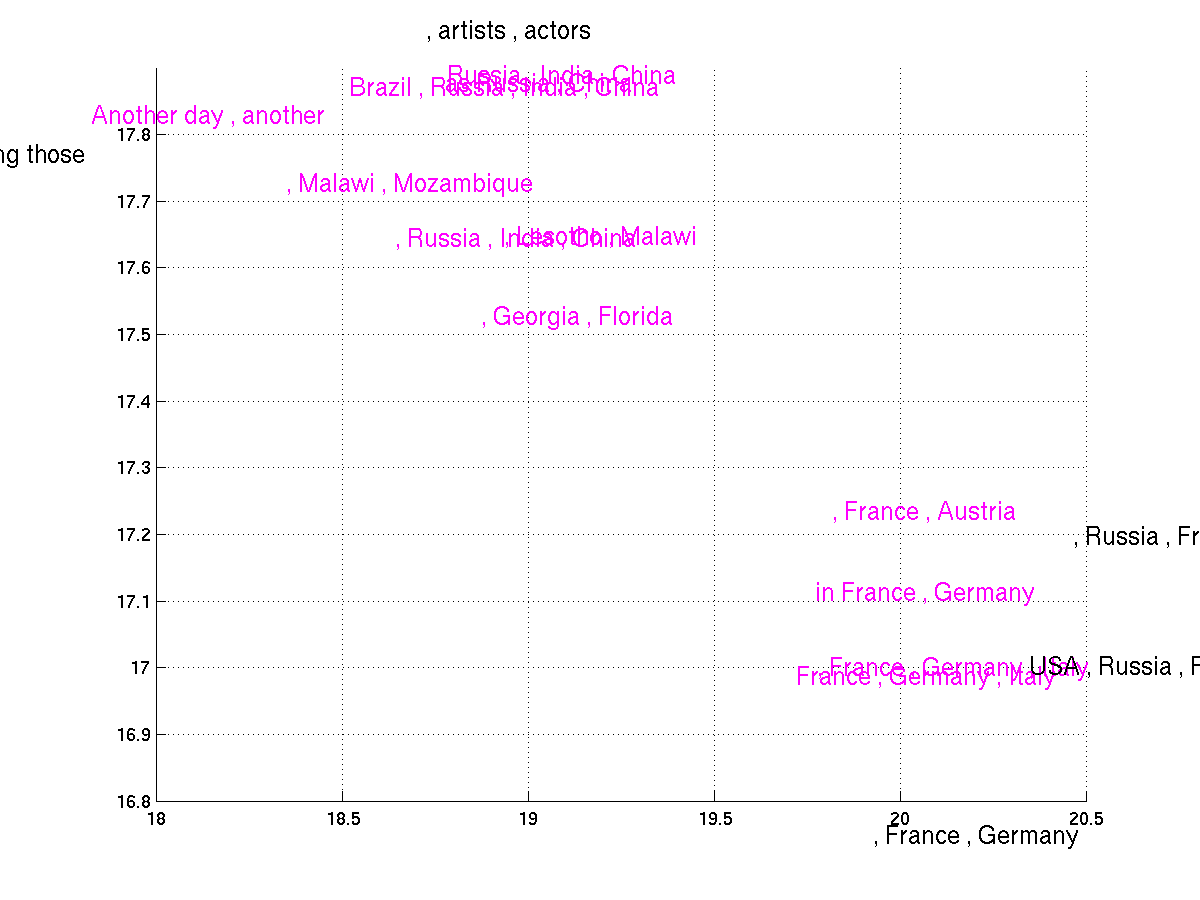}
    \end{minipage}
    \caption{2--D embedding of the learned phrase representation. The top left
    one shows the full representation space (1000 randomly selected points),
while the other three figures show the zoomed-in view of specific regions
(color--coded).} 
\end{figure*}
\end{landscape}

}

\end{document}